\documentclass{ecai}  

\usepackage{latexsym}
\usepackage{amsmath}
\usepackage{amsthm}
\usepackage{booktabs}
\usepackage{color}

\usepackage{graphicx}
\usepackage{subfigure}
\usepackage{CJKutf8}
\usepackage{extarrows}
\usepackage{url}

\usepackage{booktabs}
\usepackage{ulem}
\usepackage{bm}
\usepackage{enumitem}
\usepackage{multirow}
\usepackage{amssymb}

\begin{document}

\begin{frontmatter}

\title{LoginMEA: Local-to-Global Interaction Network for Multi-modal Entity Alignment}

\author[A,B]{\fnms{Taoyu}~\snm{Su}}
\author[A,B]{\fnms{Xinghua}~\snm{Zhang}}
\author[A]{\fnms{Jiawei}~\snm{Sheng}\thanks{Corresponding Author.}}
\author[C]{\fnms{Zhenyu}~\snm{Zhang}} 
\author[A,B]{\fnms{Tingwen}~\snm{Liu}} 

\address[A]{Institute of Information Engineering, Chinese Academy of Sciences, Beijing, China}
\address[B]{School of Cyber Security, University of Chinese Academy of Sciences, Beijing, China}
\address[C]{Baidu Inc.}
\address[]{\{sutaoyu, zhangxinghua, shengjiawei, liutingwen\}@iie.ac.cn, zhangzhenyu07@baidu.com}

\begin{abstract}
Multi-modal entity alignment (MMEA) aims to identify equivalent entities between two multi-modal knowledge graphs (MMKGs), whose entities can be associated with relational triples and related images. 
Most previous studies treat the graph structure as a special modality, and fuse different modality information with separate uni-modal encoders, neglecting valuable relational associations in modalities.
Other studies refine each uni-modal information with graph structures, but may introduce unnecessary relations in specific modalities. 
To this end, we propose a novel local-to-global interaction network for MMEA, termed as LoginMEA.
Particularly, we first fuse local multi-modal interactions to generate holistic entity semantics and then refine them with global relational interactions of entity neighbors.
In this design, the uni-modal information is fused adaptively, and can be refined with relations accordingly. 
To enrich local interactions of multi-modal entity information, we device modality weights and low-rank interactive fusion, allowing diverse impacts and element-level interactions among modalities.
To capture global interactions of graph structures, we adopt relation reflection graph attention networks, which fully capture relational associations between entities. 
Extensive experiments demonstrate superior results of our method over 5 cross-KG or bilingual benchmark datasets, indicating the effectiveness of capturing local and global interactions.
\end{abstract}

\end{frontmatter}

\section{Introduction}

\begin{figure}[t]
	\begin{center}
		\includegraphics[width=8.5cm]{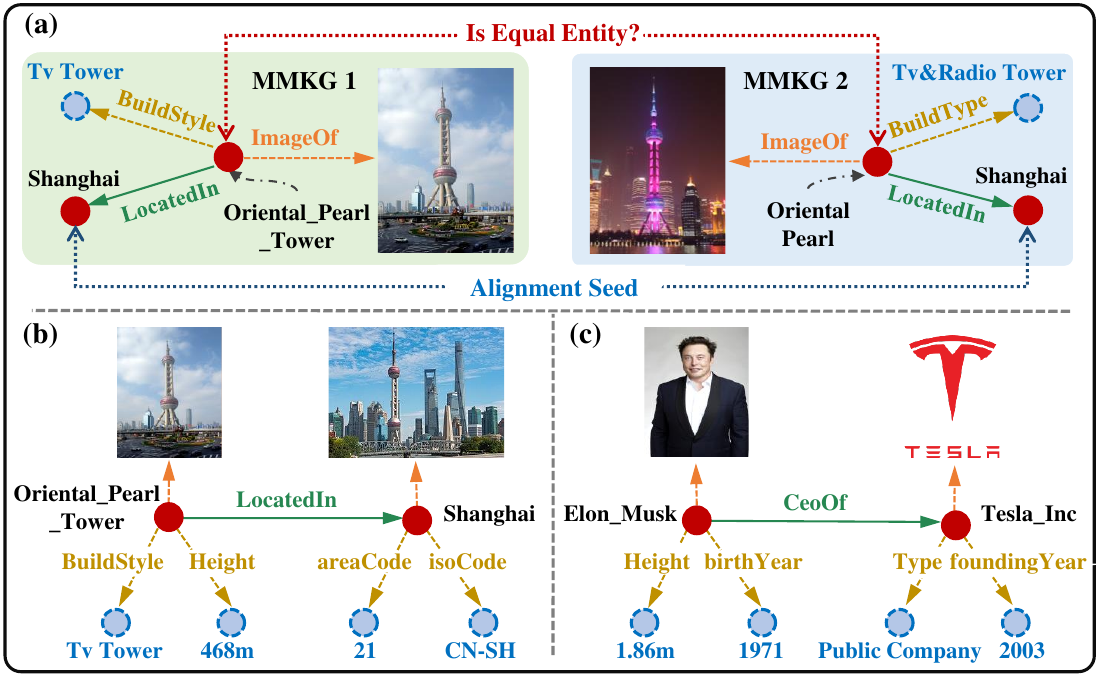}
		\caption{An example of the MMEA task between two MMKGs, where (b) and (c) showcase the relational associations of entity images in an MMKG.}
		\label{LoginMEA_Intro}
	\end{center}
\end{figure}

\textit{Knowledge graphs} (KGs) have emerged as a prominent data structure for representing factual knowledge in the form of triples, i.e., <entity, relation, entity>, where two entities are connected through relations.
\textit{Multi-modal knowledge graphs} (MMKGs) further extend traditional KGs by introducing representative multi-modal information, such as visual images, attributes, and relational data~\cite{POE}.
As a pivotal task for MMKG integration, \textit{multi-modal entity alignment} (MMEA) aims to identify equivalent entities between two MMKGs.
As shown in Figure~\ref{LoginMEA_Intro}(a), the model requires to identify entity {\small \texttt{Oriental\_Pearl\_Tower}} in \textit{MMKG-1} is equivalent to {\small \texttt{Oriental\_Pearl}} in \textit{MMKG-2}, utilizing their multi-modal information.
Such a task can benefit several downstream applications~\cite{liang2024survey}, such as recommendation systems~\cite{MKG4rRS} and question answering~\cite{MKG4QA}.

\begin{figure}[!t] 
	\begin{center}
		\includegraphics[width=8.5cm]
            {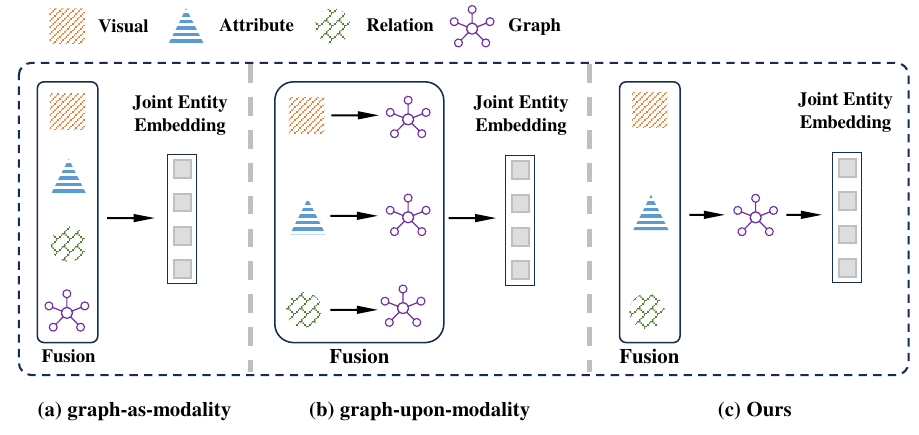}
		\caption{The schematic diagram of different modeling paradigms.}
		\label{fusion-paradigm}
	\end{center}
\end{figure}

For MMEA, a crucial problem is how to exploit the consistency of equivalent entities between different MMKGs with their multi-modal information in relational graph structures~\cite{POE}. 
To this end, most methods~\cite{MMEA,HMEA,Meaformer,XGEA} firstly encode uni-modal features, and then fuse them to obtain joint entity embedding for alignment.
They can be roughly manifested into two groups (shown in Figure~\ref{fusion-paradigm}):
\begin{itemize}[leftmargin=*]
    \item The first group of methods~\cite{EVA,MSNEA,MCLEA,Meaformer}, namely \textit{graph-as-modality}, treat the graph structure as a special modality of entities, and separately encode entity uni-modal information (i.e., structures, images, attributes, relations).
    However, these methods cannot capture valuable relational associations between entity images without graph structures.
    As shown in Figure~\ref{LoginMEA_Intro}(b), in the MMKG, entity {\small \texttt{Oriental\_Pearl\_Tower}} is located in {\small\texttt{Shanghai}}, and their images of entities reflect this {\small\texttt{LocatedIn}} relation.
    Separately encoding the uni-modal features without graph structures would neglect these beneficial relational associations, which generates suboptimal uni-modal embeddings, and impedes the final joint multi-modal entity embeddings.
    \item Another group of methods~\cite{MSNEA,XGEA}, namely \textit{graph-upon-modality}, firstly encode uni-modal information (including images, attributes and relations), and then refine these uni-modal information with graph structures. 
    However, there also exist relations that cannot be reflected in the entity images.
    As shown in Figure~\ref{LoginMEA_Intro}(c), the images of {\small \texttt{Elon\_Musk}} and {\small \texttt{Tesla\_Inc}} cannot reflect the  {\small \texttt{CeoOf}} relation.
    Directly building the relations between images would introduce unnecessary relational inductive bias~\cite{RelationalInductiveBias}, leading to unstable performance in MMEA.
\end{itemize}

Therefore, a natural idea is to fuse multi-modal information firstly to obtain holistic entity semantics, and then refine them with relational graph structures (as Figure~\ref{fusion-paradigm}(c)).
In this sense, the uni-modal information (e.g., images) is adaptively fused and accordingly refined by relations.
However, there are still two vital challenges requiring further design:
(1) \textit{How to fuse \underline{local} different modality information of an entity considering its multi-modal interactions?}
Existing multi-modal fusion mostly employs vector concatenation~\cite{POE,HMEA,MMEA} or weighted attention mechanism~\cite{EVA,MCLEA,UMAEA,PSNEA,PCMEA}.
Nevertheless, these methods only considers the importance of modality features, lacking multifarious feature interactions in entity multi-modal information.
For instance, {\small \texttt{Oriental\_Pearl\_Tower}} has interactions between visual architectural cues and attribute details (e.g., height, function), ensuring its identification as a television tower.
(2) \textit{How to build \underline{global} relational interactions between entities, while enhancing relational consistency between MMKGs?}
Conventional MMEA studies~\cite{GCN-Align, Meaformer} mostly employ vanilla graph neural networks (GNNs)~\cite{GCN,GAT} that can hardly capture relations between entity embeddings.
Relational GNNs~\cite{RGCN,liang2024mines} can be promising but may struggle in building relational consistency between different graph structures in the embedding space~\cite{RREA}.
Since structural information is reported~\cite{MMEA,Meaformer} as pivotal information for MMEA results, it requires further design for learning relational graph structures. 

To this end, we propose a novel \underline{Lo}cal-to-\underline{g}lobal \underline{i}nteraction \underline{n}etwork for \underline{M}ulti-modal \underline{E}ntity \underline{A}lignment, termed as {LoginMEA}\footnote{Our code is available at \url{https://github.com/sutaoyu/LoginMEA}, and the \textbf{Supplemental Materials} is also available there.}.
Particularly, it first fuses local multi-modal interactions to generate holistic entity semantics, then refines them with global relational interactions of entity neighbors.
To fully fuse different modality information, we propose a \textit{local multi-modal interactive fusion} module, which designs entity-specific adaptive weights and low-rank interactive fusion.
Compared to existing methods~\cite{EVA,MCLEA,PCMEA,Meaformer}, this module discerns diverse weight impacts of different entity modality information and captures multifarious element-level modality feature interactions, allowing capability for building relational associations of uni-modal information between entities. 
Besides, we propose a \textit{global multi-modal interactive aggregation} module, which adopts relational reflection graph attention networks to refine entity embeddings with entity neighbors.
Compared to vanilla GNN-based methods~\cite{GCN-Align,Meaformer}, this module fully utilizes relational interactions between entities in the MMKG, and can retain relational consistency between different MMKGs. 
Finally, we adopt a \textit{contrastive alignment loss} to train the overall model, which ensures the consistency of equivalent entities from different MMKGs to achieve the MMEA task.
The contributions of this paper are summarized as follows:
\begin{itemize}[leftmargin=*]
    \item We investigate the relational associations between entities in their multi-modal information, and we propose a novel MMEA framework, LoginMEA. To our knowledge, we are the first to build relational graph structures upon holistic entities to leverage relational associations of multi-modal information in MMEA studies.
    \item We develop the LoginMEA framework with local-to-global interaction networks, which builds multi-modal interactions of entity information with local multi-modal interactive fusion, and builds global relational interactions between joint multi-modal entity embeddings with global multi-modal interactive aggregation.
    \item Experimental results and extensive analyses confirm our significant improvements in comparison with previous state-of-the-art methods on 5 benchmark datasets.
\end{itemize}

\section{Related work} \label{Sec:related_work}

\paragraph{Entity Alignment (EA).}
Existing EA methods aim to embed entities from different KGs into a unified vector space, and identify equivalent entities by measuring the distance between their embeddings. 
Early methods~\cite{MTransE,Transedge,IPTransE,BootEA} employ TransE~\cite{TransE} or its variants to learn entity embeddings and relations.
Recognizing that entities with similar neighborhood structures are likely to be aligned, recent approaches~\cite{GCN-Align,KECG,RDGCN,MuGNN,AliNet} leverage graph neural networks (GNNs)~\cite{GCN, GAT} to capture entity structure information to enhance entity embeddings.
Besides, to compensate for limited graph structure signals in alignment learning, another line of recent studies ~\cite{JAPE,AttrE,MultiKE,KDCoE,AttrGNN} retrieve auxiliary supervision from side information such as attribute information and entity descriptions.
Although the aforementioned methods attempt to improve entity representation by utilizing structural and side information, they can hardly directly utilize visual images of entities to enhance EA in MMKGs.

\begin{figure*}[!t]
	\begin{center}
		\includegraphics[width=0.96\textwidth]
        {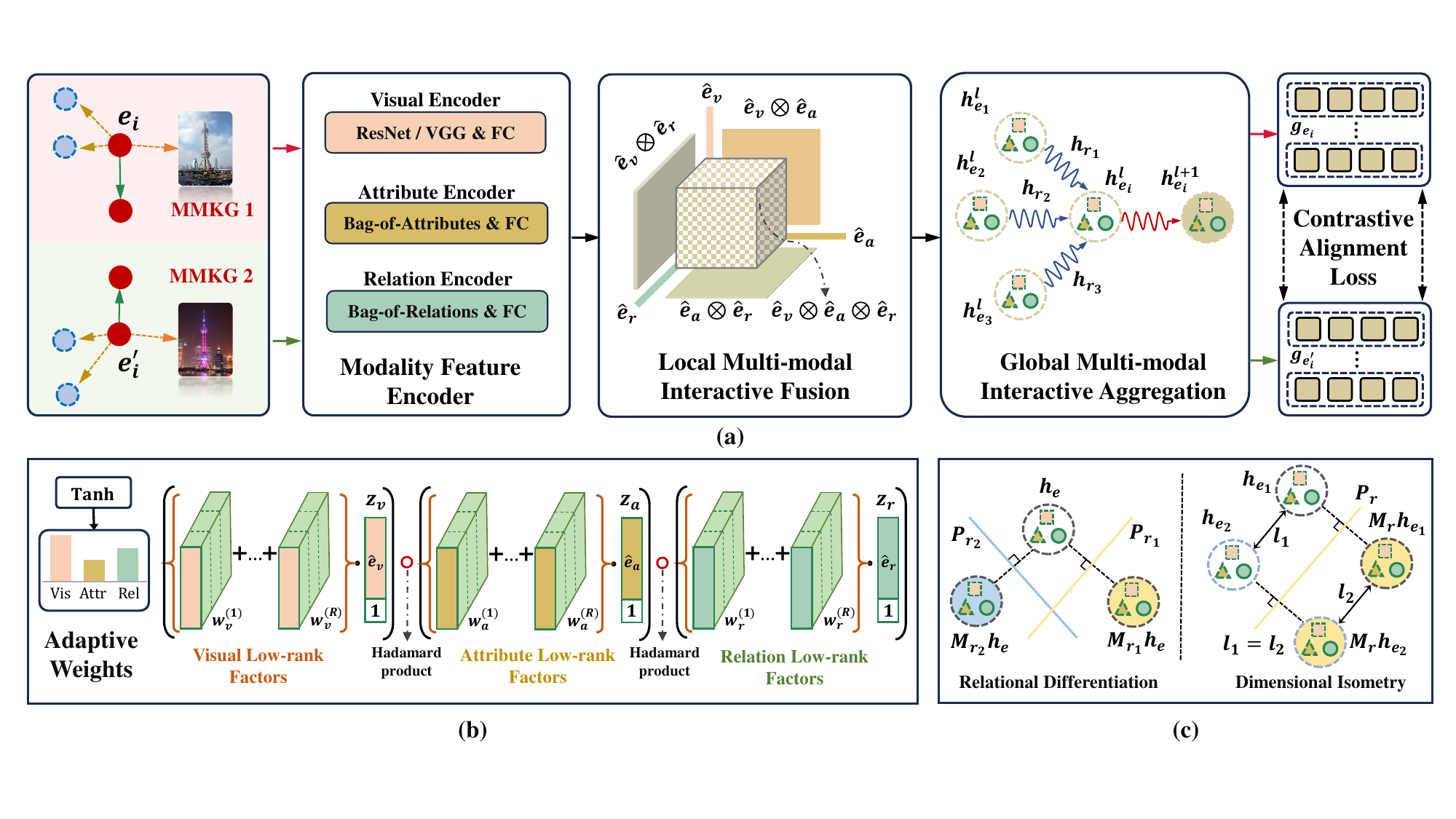}
		\caption{The LoginMEA framework for the multi-modal entity alignment, where (a) denotes the overall backbone, (b) the principle of the local multi-modal interactive fusion, (c) the principle of the global multi-modal interactive aggregation.}
		\label{LoginMEA_Main}
	\end{center}
\end{figure*}

\paragraph{Multi-Modal Entity Alignment.}

Current MMEA methods can be roughly classified into two groups according to the utilization of graph structures: 
(1) In \textit{ graph-as-modality} approaches, the graph structure is considered as a distinct modality.
Early researches~\cite{POE,HMEA,MMEA} learn modality-specific embeddings by using separate encoders, then adopt direct or fixed-weight operations to combine multi-modal information.
However, these approaches lack adaptability in learning the relative importance of different modalities.
To address this, EVA~\cite{EVA} combines multi-modal information for MMEA with learnable weighted attention to model the importance of each modality.
Building upon EVA's foundation, some subsequent works enhance entity alignment based on contrastive learning~\cite{MCLEA,UMAEA,ACK-MMEA} or generated pseudo labels~\cite{PSNEA,PCMEA}. 
Yet, these approaches fail to compute modality weights at the entity level. 
To address this limitation, recent works~\cite{Meaformer,MoAlign} leverage transformer-based approaches for multi-modal fusion.
However, the \textit{graph-as-modality} methods face limitations by treating the graph structure as a single modality, hindering the capture of relational information between entity images.
(2) In terms of \textit{graph-upon-modality} methods,  they firstly encode uni-modal information (including images, attributes, and relations), and then refine this uni-modal information with graph structures.
MSNEA~\cite{MSNEA} utilizes TransE~\cite{TransE} for uni-modal information to guide relational feature learning.
XGEA~\cite{XGEA} employs the message-passing mechanism of GNNs to aggregate uni-modal information, thereby guiding structural embedding learning.
However, learning relations directly from unimodal information that cannot reflect relations introduces unnecessary relational inductive bias.
In this work, we introduce a local-to-global interaction network, fusing local multi-modal interactions to generate holistic entity semantics and then refine them with global relational interactions of entity neighbors.

\section{Problem Formulation}
In general, a \textit{\textbf{multi-modal knowledge graph (MMKG)}} is composed of relational triples with entities, attributes and images, which can be defined as $\mathcal{G}=(\mathcal{E},\mathcal{R}, \mathcal{A}, \mathcal{V})$, where $\mathcal{E}, \mathcal{R}, \mathcal{A}, \mathcal{V}$ are the sets of entities, relations, attributes and visual images, respectively.
Therefore, the triples are defined as $\mathcal{T} \subseteq \mathcal{E}\times \mathcal{R}\times \mathcal{E}$, where each entity $e\in \mathcal{E}$ can have attributes and images.
Following previous works~\cite{EVA,MCLEA,Meaformer}, we focus on four kinds of entity information, including graph structure $g$, visual image $v$, neighboring relation $r$, and attribute $a$.

Based upon, \textit{\textbf{multi-modal entity alignment (MMEA)}}~\cite{POE,MMEA} aims to identify equivalent entities from two different MMKGs.
Formally, given two MMKGs $\mathcal{G}_{1}$ and $\mathcal{G}_{2}$ with their relational triples and multi-modal attributes, the goal of MMEA task is to identify equivalent entity pairs $\{\langle e_{1}, e_{2}\rangle|e_{1} \in \mathcal{G}_{1}, e_{2} \in \mathcal{G}_{2}, e_{1}\equiv e_{2}\}$.
For model training, a set of pre-aligned entity pairs $\mathcal{S}$ (a.k.a, \textit{alignment seed}) is provided. In the testing phase, given an entity $e\in \mathcal{G}_{1}$, the model requires to identify the equivalent $e_{2} \in \mathcal{G}_{2}$ from all candidate entities.

\section{Methodology}
In this section, we present our LoginMEA framework, illustrated in Figure~\ref{LoginMEA_Main}.
Our model consists of four modules: 1) the \textit{modality feature encoder} to generate uni-modal embeddings for each entity, 2) the \textit{local multi-modal interactive fusion} to generate joint entity embeddings with multi-modal interactions, 3) the \textit{global multi-modal interactive aggregation} to achieve relational interaction between joint multi-modal entity embeddings, and 4) the \textit{contrastive alignment loss} to achieve entity alignment for equivalent entities.

\subsection{Modality Feature Encoder} 

To capture entity features of all modalities, each modality takes a uni-modal encoder to generate the uni-modal embedding.

\paragraph{Visual Encoder.}
To capture visual features for entity images, we employ a pre-trained visual models (PVM) as the feature extractor, which generates embedded features of the visual modality through learnable convolutional layers. 
Specifically, for each image ${x}_{v}$, we input it into the PVM, and refine the output embedding with a feedforward layer, which is formulated as follows:
\begin{equation}
\begin{aligned}
\boldsymbol{e}_v = \boldsymbol{W}_{v} {\mathrm{PVM}}({x}_{v}) + b_{v}, 
\end{aligned}
\end{equation}
where $\boldsymbol{W}_{v} \in \mathbb{R}^{d \times d_v}$ and $b_{v}\in \mathbb{R}^{d}$ are learnable parameters.
Following previous works~\cite{EVA,MCLEA,Meaformer}, we adopt pre-trained VGG-16~\cite{VGG16} for cross-KG datasets (FB15K-DB15K, FB15K-YAGO15K), and ResNet-152~\cite{Resnet152} for bilingual datasets (DBP15K), respectively.

\paragraph{Attribute and Relation Encoder.}
In MMKGs, entities encompass diverse relation and attribute information.
Consequently, we follow prior approaches~\cite{EVA,MCLEA,Meaformer} and employ a method akin to a bag-of-words model to capture entity attributes and relations.
Specifically, we construct N-hot vectors for attributes and relations, where the corresponding position is set to 1 if the entity has the specific attribute or relation, otherwise 0, respectively.
Note that, following previous works~\cite{Meaformer}, we also consider the most frequently occurring top-K attributes and relations of all entities, leading to K-dimensional vectors.
Afterward, we obtain the embeddings of entity attributes and relations as follows:
\begin{equation}
\begin{aligned}
\boldsymbol{e}_{l} = \boldsymbol{W}_{l}\boldsymbol{x}_{l} + b_{l}, l \in \{a, r\},
\end{aligned}
\end{equation}
where $l \in \{a, r\}$ denotes the attribute or relation. $\bm{W}_{l}\in \mathbb{R}^{d\times d_l}$ and $b_{l}\in \mathbb{R}^d$ are learnable parameters, and $\bm{x}_{l}\in \mathbb{R}^{d_{l}}$ denotes the bag-of-attribute or bag-of-relation features, respectively. 

\subsection{Local Multi-modal Interactive Fusion}
To capture multi-modal feature interactions, we propose local multi-modal interactive fusion, allowing each entity to incorporate complementary information for holistic entity semantics. 

\paragraph{Entity-specific Adaptive Modality Weights.}
Conventional methods~\cite{POE,HMEA} simply concatenate multi-modal embeddings or adopt global modality weights for all entities, which can hardly capture diverse importance of different modalities for each entity.
Therefore, we design an entity-specific adaptive weighted mechanism for different modalities. 
For entity $e$ and its modality embedding $\boldsymbol{e}_m, m \in \{v, a, r\}$, the modality weights $\alpha_{m}$ are derived by:
\begin{equation}
\begin{aligned}
\alpha_{m}=\frac{\exp (\boldsymbol{w}_{m}^{\top} \operatorname{Tanh}(\boldsymbol{e}_m))}{\sum_{n \in \{v, a, r\}} \exp (\boldsymbol{w}_{n}^{\top} \operatorname{Tanh}(\boldsymbol{e}_n))}, 
\end{aligned}
\end{equation}
where $\boldsymbol{w}_{m}\in \mathbb{R}^{d}$ is a learnable vector of modality $m$.
According to $\alpha_{m}$ of the entity, we can control the impacts of uni-modal features in the fusion.
The weighted uni-modal embeddings is as follows:
\begin{equation}
\begin{aligned}
\boldsymbol{\hat{e}}_{m}= \alpha_{m} \boldsymbol{e}_{m}, m \in \{v, a, r\},
\end{aligned}
\end{equation}
where $\boldsymbol{\hat{e}}_{m}$ is the weighted embedding of modality $m$ of entity $e$.

\paragraph{Low-rank Interactive Fusion.}
To capture the multi-modal interactions of entity information, \textit{\textbf{tensor fusion}}~\cite{TFN} is a successful approach for multi-modal fusion, which can enrich multifarious multi-modal interactions at the embedding vector element-level. 
Specifically, the input embeddings are firstly transformed into high-dimensional tensors, and then mapped into a low-dimensional embedding space. 
The joint entity embedding $\boldsymbol{h}_e$ is derived as follows:
\begin{equation}
\begin{aligned}
\boldsymbol{h}_{e}= \boldsymbol{\mathcal{W}} \cdot \boldsymbol{\mathcal{Z}}+b,
\label{eq_tf_h}
\end{aligned}
\end{equation}
where $\boldsymbol{\mathcal{W}} \in \mathbb{R}^{(d_1 \times d_2 \times \dots \times d_M)\times d_h}$ is an ($M$+1)-order weight tensor, $b\in \mathbb{R}^{d_h}$ is the bias, and $\boldsymbol{\mathcal{Z}} \in \mathbb{R}^{d_1 \times d_2 \times \dots \times d_M}$ is a high-dimensional $M$-order tensor for the interacted multi-modal feature.
Note that the operation $\cdot$ denotes the tensor multiplication, leading to output embedding $\boldsymbol{h}_{e}\in\mathbb{R}^{d_h}$. 
Here, $\boldsymbol{\mathcal{Z}}$ is calculated by the mathematical outer product between the augmented vector of visual, attribute, and relation feature as $\boldsymbol{z}_{v} = [\mathbf{\hat{e}}_{m} \ 1]^\top$,  $\boldsymbol{z}_{a} = [\mathbf{\hat{e}}_{a} \ 1]^\top$, and $\boldsymbol{z}_{r} = [\mathbf{\hat{e}}_{r} \ 1]^\top$, respectively.
The extra constant dimension with value 1 retains the uni-modal features during the interaction, and thus $\boldsymbol{\mathcal{Z}}$ is defined as:
\begin{equation}
\begin{aligned}
\boldsymbol{\mathcal{Z}} &:= \bigotimes_{m=1}^{M} \boldsymbol{z}_{m} :=[\begin{array}{c}
\boldsymbol{\hat{e}}_{v} \\
1
\end{array}] \otimes[\begin{array}{c}
\boldsymbol{\hat{e}}_{a} \\
1
\end{array}] \otimes[\begin{array}{c}
\boldsymbol{\hat{e}}_{r} \\
1
\end{array}] \\
& =[\underbrace{\boldsymbol{\hat{e}}_{v}, \boldsymbol{\hat{e}}_{a}, \boldsymbol{\hat{e}}_{r}}_{\text{Uni\mbox{-}modal}},\underbrace{ \boldsymbol{\hat{e}}_{v} \otimes \boldsymbol{\hat{e}}_{a}, \boldsymbol{\hat{e}}_{v} \otimes \boldsymbol{\hat{e}}_{r}, \boldsymbol{\hat{e}}_{a} \otimes \boldsymbol{\hat{e}}_{r}}_{\text{Bi\mbox{-}modal}},\underbrace{\boldsymbol{\hat{e}}_{v} \otimes \boldsymbol{\hat{e}}_{a} \otimes \boldsymbol{\hat{e}}_{r}}_{\text{Tri\mbox{-}modal}}]^\top,
\label{tensor_z}
\end{aligned}
\end{equation}
where $\otimes$ denotes the outer product between vectors.
Here, we can observe that the uni-modal terms retain original information within each modality, the bi-modal terms retain interactions between two modalities, and the tri-modal terms retain interactions among three modalities.
It is worth noting that, for 3 modalities, the total number of these terms is $C_3^1 + C_3^2 + C_3^3 = 7$.
When there are $n$ modalities, it will yield $C_n^1 + C_n^2 + \cdots + C_n^n = 2^n-1$ terms. 
In this way, Eq.~(\ref{tensor_z}) enriches multifarious feature interactions at the vector element-level, and allows scalability for any number of modalities.

However, the dimensionality of the tensor $\boldsymbol{\mathcal{Z}}$ grows exponentially with the number of modalities as ${\textstyle \prod_{m=1}^{M}}d_m$. 
Additionally, the number of parameters that need to be learned in $\boldsymbol{\mathcal{W}}$ will also increase accordingly. 
This situation not only results in severe computational overhead, but may also lead to overfitting in training. 
Therefore, to alleviate this, we adopt a \textit{\textbf{low-rank multi-modal fusion}}~\cite{LMF} approach, which remains the enriched capability of feature interactions but develops an efficient manner.
Specifically, we leverage a low-rank weight decomposition to approximate original weight tensor $\boldsymbol{\mathcal{W}}$:
\begin{equation}
\begin{aligned}
\widetilde{\boldsymbol{\mathcal{W}}}:= \sum_{i=1}^{R} \bigotimes_{m=1}^{M}\boldsymbol{w}_{m}^{(i)},
\end{aligned}
\end{equation}
where $R$ is called \textit{rank} of the tensor $\widetilde{\boldsymbol{\mathcal{W}}}$, which is the minimum value that makes the decomposition valid.
There are $r$ decomposition factors with $\boldsymbol{w}_{m}^{(i)} \in \mathbb{R}^{d_m \times d_h}$, and the outer product $\bigotimes_{m=1}^{M}\boldsymbol{w}_{m}^{(i)} \in \mathbb{R}^{(d_1 \times d_2 \times \dots \times d_M)\times d_h}$.
We can easily infer that each resulted tensor of the outer product has the rank of 1, since it is linearly dependent on vectors in the tensor.
Based on the low-rank weight tensor, the tensor fusion in Eq.~(\ref{eq_tf_h}) can be derived as ($b$ is omitted here):
\begin{equation}
\begin{aligned}
\boldsymbol{h}_{e} & =(\sum_{i=1}^{R} \bigotimes_{m=1}^{M} \boldsymbol{w}_{m}^{(i)}) \cdot \boldsymbol{\mathcal{Z}} \\
& =\sum_{i=1}^{R}(\bigotimes_{m=1}^{M}\boldsymbol{w}_{m}^{(i)} \cdot \boldsymbol{\mathcal{Z}}) \\
& =\sum_{i=1}^{R}(\bigotimes_{m=1}^{M}\boldsymbol{w}_{m}^{(i)} \cdot \bigotimes_{m=1}^{M} \boldsymbol{z}_{m}) \\
& =\Lambda_{m=1}^{M}[\sum_{i=1}^{R}\boldsymbol{w}_{m}^{(i)} \cdot \boldsymbol{z}_{m}],
\label{eq_he_mid}
\end{aligned}
\end{equation}
where $\Lambda_{m=1}^{M}$ is defined as $\Lambda_{m=1}^{3} = \boldsymbol{z}_{v} \circ \boldsymbol{z}_{a} \circ \boldsymbol{z}_{r}$, and $\circ$ is Hadamard product.
Therefore, in this paper, the low-rank multi-modal fused embedding $\boldsymbol{h}_{e}$ of 3 modalities can be expressed as follows:
\begin{equation}
\begin{aligned}
\boldsymbol{h}_{e} =(\sum_{i=1}^{R} \boldsymbol{w}_{v}^{(i)} \cdot \boldsymbol{z}_{v}) \circ(\sum_{i=1}^{R} \boldsymbol{w}_{a}^{(i)} \cdot \boldsymbol{z}_{a}) \circ(\sum_{i=1}^{R} \boldsymbol{w}_{r}^{(i)} \cdot \boldsymbol{z}_{r}),
\label{eq_he_final}
\end{aligned}
\end{equation}
which enables to derive $\boldsymbol{h}_{e}$ directly based on the uni-modal embeddings and modal-specific decomposition factors, avoiding the heavy computation of large input tensor $\boldsymbol{\mathcal{Z}}$ and weight tensor $\boldsymbol{\mathcal{W}}$ in Eq.~(\ref{eq_tf_h}), while still allowing element-level multi-modal interactions.
\paragraph{Remarks.}
Actually, Eq.~(\ref{eq_he_mid}) reduces the computational complexity of tensorization and fusion from $O(d_h \times {\textstyle \prod_{m=1}^{M}d_m})$ to $O(d_h \times r \times {\textstyle \sum_{m=1}^{M}d_m})$, and adopts less parameters to avoid overfitting.
Besides, Eq.~(\ref{eq_he_mid}) comprises fully differentiable operations, allowing the parameters $\boldsymbol{w}_{m}^{(i)}$ to be learned via back-propagation.
Moreover, as $\boldsymbol{z}_{m}$ involves entity-specific weighted modality embedding $\boldsymbol{\hat{e}}_{m}$, the final fused multi-modal embedding $\boldsymbol{h}_{e}$ not only encompasses multifarious inter-modality interaction details as in Eq.~(\ref{tensor_z}), but also captures entity-specific importance of different modalities of each entity.

\subsection{Global Multi-modal Interactive Aggregation} 
To perceive structural information of entities, we aggregate entity neighbors based on the holistic joint multi-modal entity embeddings, retaining relational graph structures for better entity embeddings.

\paragraph{Relational Reflection Graph Attention Network.}
To capture different importance of entity neighbors, it is intuitive to employ graph attention networks (GATs)~\cite{GAT}. 
However, vanilla GAT can hardly capture diverse relations between entities.
To this end, we adopt a relational reflection graph attention network~\cite{RREA} to aggregate entity neighbors retaining relational structural information. 
Specifically, the $l$-th layer's embedding for entity $e_i$ can be obtained as follow:
\begin{equation}
\begin{aligned}
\boldsymbol{h}_{e_{i}}^{l+1}=\operatorname{Tanh}\left(\sum_{r_{j}, e_{j} \in \mathcal{N}(e_{i})} \phi(r_j) \boldsymbol{M}_{r_j} \boldsymbol{h}_{e_{j}}^{l}\right),
\end{aligned}
\end{equation}
where $\mathcal{N}(e_{i})$ denotes the set of neighboring relations and entities. $e_j$ and $r_{j}$ denotes the neighboring entity and relation, respectively. $\phi(r_{j})$ is a relation-specific scalar, controlling the importance of relation $r_{j}$ in aggregating the corresponding neighboring entities.
Here, $\boldsymbol{M}_{r_{j}}$ is a relational transformation matrix reflecting relation $r_{j}$, which naturally ensures the same entity is transformed by different relations distinguishable in different positions, namely \textit{relational differentiation} property~\cite{RREA} (as shown in Figure~\ref{LoginMEA_Main}(c)).

However, it is reported in~\cite{RREA} that a transformation matrix without constraints can hardly remain the \textit{dimensional isometry} property~\cite{RREA}, i.e., \textit{when two entity embeddings are transformed by the same relation, their norms and relative distance should be retained} (as shown in Figure~\ref{LoginMEA_Main}(c)).
Therefore, if two entities from different MMKGs are aligned, their neighbors with the same relation can be easily aligned in the embedding space.
To remain this relational consistency, a simple yet effective way is to constrain $\boldsymbol{M}_{r_j}$ as an orthogonal matrix. 
We refer the readers to the literature~\cite{RREA}.
For implementation, $\boldsymbol{M}_{r_j}$ can be achieved by:
\begin{equation}
\begin{aligned}
\boldsymbol{M}_{r_j}:=\bm{I}-2 \bm{h}_{r_j} \bm{h}_{r_j}^{T},
\end{aligned}
\end{equation}
where $\bm{I}$ is the identity matrix, and $\bm{h}_{r_{j}}\in \mathbb{R}^{d}$ denotes the learnable relation embedding of $r_j$.
Here $\bm{h}_{r_{j}}$ is randomly initialized, and keeps normalized in learning to ensure $\left\|\bm{h}_{r_{j}}\right\|_{2}=1$. 
The proof for orthogonality is shown in Eq.~(\ref{eq:proof}).
Using the relation embeddings, we can easily define the importance of neighbors with different relations.
Similar to GAT~\cite{GAT}, we achieve $\phi(r_j)$ by:
\begin{equation}
\begin{aligned}
\phi(r_j)=\frac{\exp (\boldsymbol{q}^{T}\boldsymbol{h}_{r_j})}{\sum_{r_k, e_{k} \in \mathcal{N}(e_{i})} \exp (\boldsymbol{q}^{T}\boldsymbol{h}_{r_k}))},
\end{aligned}
\end{equation}
where $\boldsymbol{q}\in \mathbb{R}^d$ denotes a learnable vector to measure the importance of the relation.
To perceive global multi-hop structures, following previous studies~\cite{GAT,AliNet}, we collect multi-hop neighboring information by stacking entity embeddings from different layers:
\begin{equation}
\begin{aligned}
\boldsymbol{g}_{e_{i}}=\left[\boldsymbol{h}_{e_{i}}^{0}\| \boldsymbol{h}_{e_{i}}^{1}\|\ldots\| \boldsymbol{h}_{e_{i}}^{l}\right],
\end{aligned}
\end{equation}
where $\|$ denotes concatenation.
Note that, at the first layer, we initialize the input embeddings $\boldsymbol{h}_{e_i}^0$ with locally fused multi-modal embeddings, and then perceive global information with stacked layers.

\paragraph{Remarks.}
It is easy to prove that $\boldsymbol{M}_{r_{j}}$ is an orthogonal matrix with constraint $\left\|\bm{h}_{r_{j}}\right\|_{2}=1$, which can be derived by:
\begin{equation}
\begin{aligned}
\boldsymbol{M}_{r_{j}}^{T} \boldsymbol{M}_{r_{j}} & = (\bm{I}-2 \bm{h}_{r_{j}}
\bm{h}_{r_{j}}^{T})^{T}(\bm{I}-2 \bm{h}_{r_{j}}  
\bm{h}_{r_{j}}^{T}) \\
& = \bm{I}-4 \bm{h}_{r_{j}} \bm{h}_{r_{j}}^{T}+4 \bm{h}_{r_{j}} \bm{h}_{r_{j}}^{T} \bm{h}_{r_{j}} \bm{h}_{r_{j}}^{T} = \bm{I}.
\end{aligned}\label{eq:proof}
\end{equation}
The number of parameters of $\boldsymbol{M}_{r_j}$ is $|\mathcal{R}|\times d$ much less than $|\mathcal{R}|\times d^2$ in vanilla transformation matrix~\cite{RGCN}.
In this way, we not only build relational structures between multi-modal entity information, but also retain the relational consistency between differnt MMKGs.

\subsection{Contrastive Alignment Loss}

To ensure the consistency of equivalent entities for MMEA, inspired by contrastive learning works~\cite{MoCO, InfoNCE}, we define the training loss as:
\begin{equation}
\begin{aligned}
\mathcal{L}= \sum_{(e_i,e_j) \in \mathcal{S}} -\log \frac{\exp (\operatorname{sim}(\boldsymbol{g}_{e_{i}}, \boldsymbol{g}_{e_{j}}) / \tau)}{ \sum_{(e_i,e_k) \notin \mathcal{S}} \exp (\operatorname{sim}(\boldsymbol{g}_{e_{i}}, \boldsymbol{g}_{e_{k}})) / \tau)},
\end{aligned}
\end{equation}
where $\mathcal{S}$ denotes the set of pre-aligned entity pairs, served as positive samples.
For each positive entity pair, we create $K$ negative entity pairs by replacing $e_j\in \mathcal{G}_2$ with false entity $e_k\in \mathcal{G}_2$.  
$\tau$ is a temperature factor, where a smaller $\tau$ emphasizes more on hard negatives, and we achieve $\operatorname{sim}(\cdot)$ with cosine similarity for simplicity.

\section{Experiments}

\subsection{Experimental Settings}

\paragraph{Datasets.}
Following prior studies~\cite{MCLEA,Meaformer}, we employ two types of multi-modal entity alignment (EA) datasets. (1) Cross-KG datasets: we select FB15K-DB15K and FB15K-YAGO15K public datasets, which are deemed as the most typical datasets for MMEA task~\cite{MMEA}. 
(2) Bilingual datasets: DBP15K~\cite{JAPE,EVA} is a commonly used benchmark for bilingual entity alignment, which contains three datasets built from the multilingual versions of DBpedia, including DBP15K$_{\rm ZH\mbox{-}EN}$, DBP15K$_{\rm JA\mbox{-}EN}$ and DBP15K$_{\rm FR\mbox{-}EN}$. 
Each of the bilingual datasets contains about 400K triples and 15K pre-aligned entity pairs.
We show the dataset details in \textbf{supplementary materials}. 
Notably, there are fewer relations, attributes and images in YAGO15K, which may lead to a sparser graph and the greater alignment difficulty.
Following previous works~\cite{MCLEA,Meaformer}, 
we utilize 20$\%$, 50$\%$, 80$\%$ of true entity pairs as alignment seeds for training on cross-KG datasets, whereas we use 30$\%$ of entity pairs as alignment seeds for training on bilingual datasets.
For the entities without corresponding images, we assign random vectors for the visual modality, as the setting of previous methods~\cite{MCLEA,Meaformer}.
\paragraph{Evaluation Metrics.}
In adherence to prior works ~\cite{EVA,MCLEA,Meaformer}, evaluation metrics utilized include Hits@1 (H@1), Hits@10 (H@10), and Mean Reciprocal Rank (MRR). 
Hits@N denotes the proportion of correct entities ranked in the top-N ranks, while MRR is the average reciprocal rank of correct entities. 
Higher values of Hits@N and MRR indicate better performance.

\begin{table*}[!th]
\scriptsize
\centering
\caption{Experimental results on the 2 cross-KG datasets, including \textbf{FB15K-DB15K} (\textbf{FB-DB15K} for short) and \textbf{FB15K-YAGO15K} (\textbf{FB-YG15K} for short). We evaluate our model in different resource settings, with 20\%, 50\% and 80\% seed alignments as in previous studies~\cite{Meaformer,MCLEA}. The best result is \textbf{bold-faced} and the runner-up is \underline{underlined}. $*$ indicates that the results are reproduced by the official source code.}
\label{fbdb_result}
\renewcommand{\arraystretch}{1.4} 
\setlength\tabcolsep{4.0pt}
{
\begin{tabular}{@{}lccccccccc|ccccccccc@{}}
\toprule
\multirow{2.5}{*}{\bf Methods} & \multicolumn{3}{c}{\bf FB-DB15K (20\%)} & \multicolumn{3}{c}{\bf FB-DB15K (50\%)} &  \multicolumn{3}{c|}{\bf FB-DB15K (80\%)} & \multicolumn{3}{c}{\bf FB-YG15K (20\%)} & \multicolumn{3}{c}{\bf FB-YG15K ((50\%)} &  \multicolumn{3}{c}{\bf FB-YG15K (80\%)}   \\

\cmidrule(r){2-4}\cmidrule(r){5-7}\cmidrule(r){8-10} \cmidrule(r){11-13}\cmidrule(r){14-16}\cmidrule(r){17-19}
& H@1  & H@10   &  MRR  
& H@1  & H@10   &  MRR  
& H@1  & H@10   &  MRR     
& H@1  & H@10   &  MRR     
& H@1  & H@10   &  MRR     
& H@1  & H@10   &  MRR     
\\

\midrule
TransE~\cite{TransE}  
& .078  & .240   &  .134   
& .230 & .446   & .306  
& .426 & .659   & .507 
& .064 & .203   &  .112   
& .197 & .382   &  .262  
& .392 & .595   &  .463     \\

IPTransE~\cite{IPTransE}  
& .065 & .215   &  .094   
& .210 & .421   &  .283  
& .403 & .627   &  .469     
& .047  & .169   & .084  
& .201  &  .369   & .248  
& .401 & .602   & .458    \\

GCN-align~\cite{GCN-Align}  
& .053 &  .174   &  .087   
& .226 &  .435   &  .293  
& .414 &  .635   &  .472     
& .081  & .235   & .153   
& .235  & .424   & .294  
& .406  & .643   & .477     \\

KECG*~\cite{KECG}  
& .128 &  .340   &  .200   
& .167 &  .416   &  .251  
& .235 &  .532   &  .336     
& .094   & .274   &  .154   
& .167  & .381   & .241  
& .241  & .501   & .329     \\

\midrule

POE~\cite{POE} 
& .126 &  .151   &  .170   
& .464 &  .658   &  .533  
& .666 &  .820   &  .721     
& .113 &  .229   & .154   
& .347 &  .536   & .414  
& .573 &  .746   & .635     \\

Chen et al.~\cite{MMEA} 
& .265 &  .541   &  .357   
& .417 &  .703   &  .512  
& .590 &  .869   &  .685     
& .234 &  .480   &  .317   
& .403 &  .645   &  .486  
& .598 &  .839   &  .682     \\

HMEA~\cite{HMEA}
& .127 &  .369   &  -   
& .262 &  .581   &  -  
& .417 &  .786   &  -     
& .105 &  .313   &  -   
& .265 &  .581   &  -  
& .433 &  .801   &  -     \\

EVA~\cite{EVA}  
& .134 &  .338   &  .201   
& .223 &  .471   &  .307  
& .370 &  .585   &  .444     
& .098 &  .276   &  .158   
& .240 &  .477   &  .321  
& .394 &  .613   &  .471     \\

MSNEA~\cite{MSNEA}
& .114 &  .296   &  .175   
& .288 &  .590   &  .388  
& .518 &  .779   &  .613     
& .103 &  .249   &  .153   
& .320 &  .589   &  .413  
& .531 &  .778   &  .620     \\

ACK-MMEA~\cite{ACK-MMEA}  
& .304 &  .549   &  .387   
& .560 &  .736   &  .624  
& .682 &  .874   &  .752     
& .289 &  .496   &  .360   
& .535 &  .699   &  .593
& .676 &  .864   &  .744     \\

XGEA*~\cite{XGEA}  
& .475 &  .739   &  .565   
& .681 &  .857   &  .746  
& .791 &  .919   &  .840     
& .431 &  .691   &  .521   
& .585 &  .801   &  .666
& .705 &  .873   &  .768     \\

UMAEA*~\cite{UMAEA}
& .533 &  \underline{.813}   &  .633   
& .664 &  .868   &  .740
& \underline{.817} &  .915   &  \underline{.853}    
& .422 &  \underline{.695}   &  .520
& .599 &  .783   &  .668 
& \underline{.728} &  .862   &  .778     \\

MCLEA~\cite{MCLEA}  
& .445 &  .705   &  .534   
& .573 &  .800   &  .652
& .730 &  .883   &  .784     
& .388 &  .641   &  .474
& .543 &  .759   &  .616 
& .653 &  .835   &  .715     \\

MEAformer~\cite{Meaformer}  
& \underline{.578} & .812   &  \underline{.661}   
& \underline{.690} & \underline{.871}   &  \underline{.755}
& .784 & \underline{.921}   &  .834    
& \underline{.444} &  .692   &  \underline{.529}   
& \underline{.612} &  \underline{.808}   &  \underline{.682}
& .724 &  \underline{.880}   &  \underline{.783}     \\

\midrule
LoginMEA (Ours)
    & \textbf{.667} & \textbf{.854}   & \textbf{.735}   
    & \textbf{.758} & \textbf{.898}   & \textbf{.810} 
    & \textbf{.843} & \textbf{.942}   & \textbf{.880}       
    & \textbf{.622}   &  \textbf{.818}   &  \textbf{.691}  
    & \textbf{.706}    &  \textbf{.865}   &  \textbf{.763}  
    & \textbf{.780}   &  \textbf{.933}   &  \textbf{.833}     \\
\bottomrule

\end{tabular}
}
\end{table*}

\begin{table}[!t]
    \scriptsize
    \centering
    \caption{Experimental results on 3 bilingual datasets, including \textbf{DBP15K}$_{\rm ZH\mbox{-}EN}$, \textbf{DBP15K}$_{\rm JA\mbox{-}EN}$ and \textbf{DBP15K}$_{\rm FR\mbox{-}EN}$. The best result is \textbf{bold-faced} and the runner-up is \underline{underlined}. $*$ indicates the results are reproduced by the official source code.}
    \label{fbdbp_result}
    \renewcommand{\arraystretch}{1.4} 
    \setlength\tabcolsep{2.0pt}{
    \begin{tabular}{@{}lccccccccc@{}}
    \toprule
    \multirow{2.5}{*}{\bf Methods} & \multicolumn{3}{c}{\textbf{DBP15K}$_{\rm ZH\mbox{-}EN}$} & \multicolumn{3}{c}{\textbf{DBP15K}$_{\rm JA\mbox{-}EN}$} &  \multicolumn{3}{c}{\textbf{DBP15K}$_{\rm FR\mbox{-}EN}$}    \\
    \cmidrule(r){2-4}\cmidrule(r){5-7}\cmidrule{8-10}
    & H@1  & H@10   &  MRR  
    & H@1  & H@10   &  MRR  
    & H@1  & H@10   &  MRR 
    \\

    \midrule
    
    GCN-align~\cite{GCN-Align}   
    & .434 &  .762  &  .550   
    & .427 &  .762  &  .540  
    & .411 &  .772  &  .530    \\
    
    KECG~\cite{KECG}    
    & .478 &  .835   &  .598   
    & .490 &  .844   &  .610  
    & .486 &  .851   &  .610     \\

    BootEA~\cite{BootEA}  
    & .629 &  .847  &  .703   
    & .622 &  .854   & .701  
    & .653 &  .874   & .731     \\

    NAEA~\cite{NAEA}  
    & .650 &  .867  & .720   
    & .641 &  .873  & .718  
    & .673 &  .894  & .752     \\
    
    \midrule

    EVA~\cite{EVA}   
    & .761 &  .907   &  .814  
    & .762 &  .913   &  .817  
    & .793 &  .942   &  .847     \\
    
    MSNEA~\cite{MSNEA} 
    & .643 &  .865   &  .719  
    & .572 &  .832   &  .660  
    & .584 &  .841   &  .671     \\

    XGEA*~\cite{XGEA}    
    & .803 &  .939   &  .854
    & .794 &  .942   &  .849 
    & .821 &  .954   &  .871     \\    

    UMAEA*~\cite{UMAEA}    
    & .811 &  .969   &  .871
    & .812 &  .973   &  .873 
    & .822 &  .981   &  .884     \\
    
    PSNEA~\cite{PSNEA} 
    & .816 &  .957   &  .869
    & .819 &  .963   &  .868 
    & .844 &  \underline{.982}   &  .891     \\

    MCLEA~\cite{MCLEA}   
    & .816 &  .948   &  .865
    & .812 &  .952   &  .865 
    & .834 &  .975   &  .885     \\
    
    MEAformer~\cite{Meaformer}     
    & \underline{.847} &  \underline{.970}   &  \underline{.892}   
    & \underline{.842} &  \underline{.974}   &  \underline{.892}
    & \underline{.845} &  .976   &  \underline{.894}     \\
    \midrule
    LoginMEA (Ours)  
    & \textbf{.873}   &  \textbf{.978}   &  \textbf{.913}  
    & \textbf{.866}    &  \textbf{.981}   &  \textbf{.911}  
    & \textbf{.881}   &  \textbf{.988}   &  \textbf{.924}     \\
    
\bottomrule

\end{tabular}
}
\end{table}

\paragraph{Baselines.}
We compare the proposed LoginMEA with the following competitive entity alignment baselines, including two groups:
\textbf{Traditional EA Methods:}
(1) \textbf{TransE}~\cite{TransE} assumes that the entity embedding ought to closely align with the sum of the attribute embedding and their relation.
(2) \textbf{IPTransE}~\cite{IPTransE} introduces an iterative entity alignment mechanism, employing joint knowledge embeddings to encode entities and relations across multiple KGs into a unified semantic space.
(3) \textbf{GCN-align}~\cite{GCN-Align} utilizes Graph Convolutional Networks~\cite{GCN} to generate entity embeddings and combines them with attribute embeddings to align entities.
(4) \textbf{KECG}~\cite{KECG} proposes a semi-supervised entity alignment method through joint knowledge embedding and cross-graph model learning.
\textbf{Multi-modal EA Methods:}
(1) \textbf{POE}~\cite{POE} defines overall probability distribution as the product of all uni-modal experts.
(2) \textbf{Chen et.al}~\cite{MMEA} designs a multi-modal fusion module to integrate knowledge representations from multiple modalities.
(3) \textbf{HMEA}~\cite{HMEA} combines the structure and visual representations in the hyperbolic space.
(4) \textbf{EVA}~\cite{EVA} integrates multi-modal information into a joint embedding, empowering the alignment model to auto-optimize modality weights.
(5) \textbf{MSNEA}~\cite{MSNEA} develops a vision-guided relation learning mechanism for inter-modal knowledge enhancement.
(6) \textbf{ACK-MMEA}~\cite{ACK-MMEA} designs a multi-modal attribute uniformization method to 
generate an attribute-consistent MMKG.
(7) \textbf{XGEA}~\cite{XGEA} proposes a structural-visual attention network to guide the learning of embeddings.
(8) \textbf{UMAEA}~\cite{UMAEA} introduces multi-scale modality hybrid for modality noise.
(9) \textbf{PSNEA}~\cite{PSNEA} advocates an incremental alignment pool strategy to alleviate alignment seed scarcity issues.
(10) \textbf{MCLEA}~\cite{MCLEA} performs contrastive learning to jointly model intra-modal and inter-modal interactions in MMKGs.
(11) \textbf{MEAformer}~\cite{Meaformer} utilizes a transformer-based fusion method to predict relatively mutual weights among modalities for each entity.

Among the methods, MCLEA and MEAformer are typical competitive methods, where MCLEA enhances single-modal representation relevance via contrastive learning, 
while MEAformer develops transformer-based attention multi-modal fusion method. 

\paragraph{Implementation Details.}
In our experiments, the graph encoder is configured with a hidden layer size of $d_g=300$ across 3 layers. 
The visual feature dimension $d_v$ is allocated 4096, while the attribute and relation feature sizes $d_a$ and $d_r$ are configured at 1000. 
The graph embedding output is fixed at a size of 300, whereas the embedding sizes for other modalities are determined to be 100. 
The training was conducted over 600 epochs with a batch size of 3,500. 
AdamW optimizer~\cite{AdamW} was employed with a learning rate of 5e-3 and a weight decay of 1e-2. 
Following previous works~\cite{EVA,MCLEA,Meaformer}, we adopt an iterative training strategy to overcome the lack of training data in the same way, and also do not consider entity names for fair comparison.
All the experiments are conducted on a 64-bit machine with two NVIDIA A100 GPUs, and 256 GB RAM memory.
Our best hyper-parameters are reported in the \textbf{supplementary materials}, which are tuned by grid search according to MRR metric.

\subsection{Overall Results}
To verify the effectiveness of LoginMEA, we report overall average results on cross-KG and bilingual datasets as shown in Table~\ref{fbdb_result} and Table~\ref{fbdbp_result}, respectively.
From the tables, we have several observations: 
(1) \textit{Our proposed method outperforms all compared baseline models on 9 benchmarks in terms of three key metrics (H@1, H@10, and MRR). }
Specifically, our model consistently outperforms state-of-the-art (SOTA) baselines, achieving significant improvements in Hits@1 scores across ${\rm ZH\mbox{-}EN/JA\mbox{-}EN/FR\mbox{-}EN}$ datasets with DBP15K, and elevates the existing high-performing Hits@1 scores from .847/.842/.845 to .873/.866/.881.
Moreover, our model brings about an average increase of 3.9\% and 8.1\% in H@1 on cross-KG datasets at 80\% and 50\% seed settings, respectively.
(2) \textit{Our model achieves better results in relatively low-resource data scenario.}
Compared to the runner-up method results, our model achieves an even more pronounced average gain of 13.3\% in H@1 and 11.8\% in MRR on cross-KG datasets with a 20$\%$ alignment seed setting. 
Our local multi-modal interacted fusion module enhances expressive entity embeddings in such scenarios by facilitating multi-modal information deep interaction.
(3) \textit{Compared to traditional EA models, the MMEA models all show significant enhancements.}
Remarkably, our model exhibits substantial enhancements in H@1 scores, with an average increase of 47.6\% (ranging from 37.4\% to 53.9\%) on Cross-KG datasets and an average improvement of 21.8\% (ranging from 20.8\% to 22.5\%) on cross-lingual datasets.
This demonstrates a significant improvement in entity alignment through the incorporation of multi-modal information.
All the results demonstrate the effectiveness of our proposed LoginMEA model.

\subsection{Ablation Study}
To investigate the impact of each module in LoginMEA, we design two groups of variants in the ablation study:
(1) LoginMEA with various components, such as removing or replacing specific modules.
(2) LoginMEA without one specific modality, including visual, relation, and attribute.
We conduct variant experiments on two Cross-KG datasets with 20\% alignments seeds, showcasing in Table~\ref{ablation_result}.

From the first group of variants, we remove the low-rank module and adaptive weights from the local interaction fusion module, causing a decline in performance.
Notably, the absence of the low-rank module has a greater impact, highlighting the importance of effective inter-modality interaction over modality weight information. 
Furthermore, replacing the local interaction fusion module with a simple concatenation fusion module led to a more significant drop, confirming its effectiveness. 
It dramatically degrades the performance when replacing the global interaction aggregation module with GAT, which emphasizes its crucial role in learning relational global multi-modal information interaction. 
Lastly, substituting our alignment contrastive loss with Intra-modal Contrastive Loss (ICL) used in previous studies~\cite{MCLEA,Meaformer}, resulted in a decrease in overall performance, validating the effectiveness of our original loss function.

From the second group of variants, we observe varying degrees of performance decline upon removing different modalities. 
The removal of any modality information affects our model's local multi-modal interaction of the fusion module.
Notably, we notice that the removal of relations has a relatively greater impact on overall performance compared to visual and attribute modalities. 
This can be attributed to the high frequency and importance of relations in Cross-KG knowledge graphs, influencing the global multi-modal information interaction within our aggregation module.

\begin{table}[!t]
    \scriptsize
    \centering
    \caption{Variant experiments on \textbf{FB15K-DB15K} and \textbf{FB15K-YG15K} (20\%). 
    ``\textit{w/o}'' means removing corresponding module from the complete model. ``repl.'' means replacing corresponding module with the other module.}
    \label{ablation_result}
        \renewcommand{\arraystretch}{1.5} 
        \setlength\tabcolsep{5pt}
        {
        \begin{tabular}{@{}rlcccccc@{}}
		
        \toprule

       &  \multirow{2.5}{*}{Model} 
       &  \multicolumn{3}{c}{FB15K-DB15K}  & \multicolumn{3}{c}{FB15K-YG15K}  \\ 

        \cmidrule(r){3-5} \cmidrule(r){6-8} 
        &
        & H@1  & H@10   &  MRR  
        & H@1  & H@10   &  MRR 
        \\
        
        \midrule
        \multirow{5}{*}[-4.5ex]{\rotatebox{90}{\textbf{Component}}}
        & LoginMEA  
        & \textbf{.667} & \textbf{.854}   & \textbf{.735}  
        & \textbf{.622}   &  \textbf{.818}   &  \textbf{.691} \\
        
        \cmidrule{1-8}

        & \textit{w/o} Low-rank  
        & .607    & .816  & .683           
        & .563    & .763   & .633  \\

        & \textit{w/o} Adaptive weights   
        & .639  & .848  & .714 
        & .606    & .817  & .681   \\

        & \textit{repl.} Concate Fusion
        & .517    & .757   & .603   
        & .513    & .735   & .591  \\

        & repl. GAT
        & .474    & .671   & .542     
        & .349    & .546   & .416  \\
        
        & repl. ICL
        & .605    & .835   & .693   
        & .542    & .785   & .628  \\

        \midrule
        \multirow{5}{*}[2.5ex]{\rotatebox{90}{\textbf{Modality}}}
        & \textit{w/o} Visual
        & .629   & .847  & .712 
        & .595    & .804  & .670   \\

        & \textit{w/o} Attribute
        & .634  & .832  & .706   
        & .579  & .788  & .653      \\

        & \textit{w/o} Relation  
        & .612   & .830   & .692   
        & .580    & .799   & .657   \\
        \bottomrule
        \end{tabular}
	}
\end{table}

\subsection{Performance under Different Modeling Paradigms}
To further validate the effectiveness of our proposed modeling paradigm for MMEA task, we implement two variants according to the graph-as-modality and graph-upon-modality paradigm mentioned in Figure~\ref{fusion-paradigm} based on the modules of LoginMEA:
(1) \textbf{LoginMEA-GAM} is implemented according to the \textit{graph-as-modality} paradigm, where the modality feature encoders, the structural modeling of graph, and the fusion module are all consistent with our LoginMEA method.
(2) \textbf{LoginMEA-GUM} follows the \textit{graph-upon-modality} paradigm, where all basic modules also maintain the same with LoginMEA to accurately explore the impact of paradigm that refines entities with relational structures on all modalities.

\begin{figure}[!t] 
	\begin{center}
		\includegraphics[width=8.5cm]
            {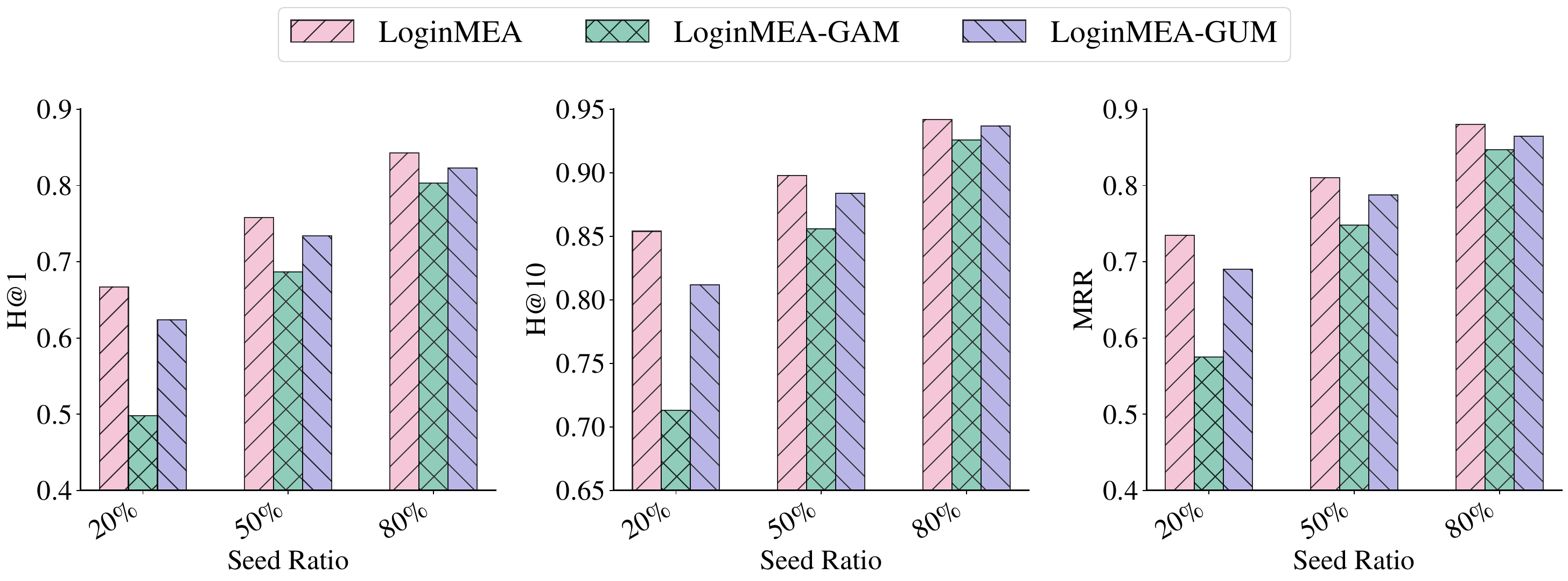}
		\caption{Results of different modeling paradigms for MMEA task on FB15K-DB15K with different seed ratios.}
		\label{Diff-Fusion-res}
	\end{center}
\end{figure}

For LoginMEA, LoginMEA-GAM and LoginMEA-GUM, we conduct experiments under various alignment seed settings on FB15K-DB15K dataset, with results depicted in Figure~\ref{Diff-Fusion-res}.
We can observe that our LoginMEA consistently achieves the best performance across different alignment seeds and metrics. 
This confirms the efficacy of our proposed paradigm, which first involves a local interactive fusion for more precise and holistic entity representations, and then follows by a global interactive aggregation upon the graph structure.
Furthermore, compared with LoginMEA-GAM, LoginMEA-GUM shows better performance due to its full aggregation of all multi-modal information upon relational structures, which facilitates the learning of relational associations in modalities.
However, the absence of local interactions among entity modalities leads to sub-optimal results of LoginMEA-GUM compared to LoginMEA.
Additionally, LoginMEA significantly outperforms LoginMEA-GAM and LoginMEA-GUM under the 20\% alignment seed setting, confirming the obvious advantage of the modeling paradigm in LoginMEA that enhances the distinctiveness of entity embeddings.

\begin{figure}[!t] 
	\begin{center}
		\includegraphics[width=8.5cm]
            {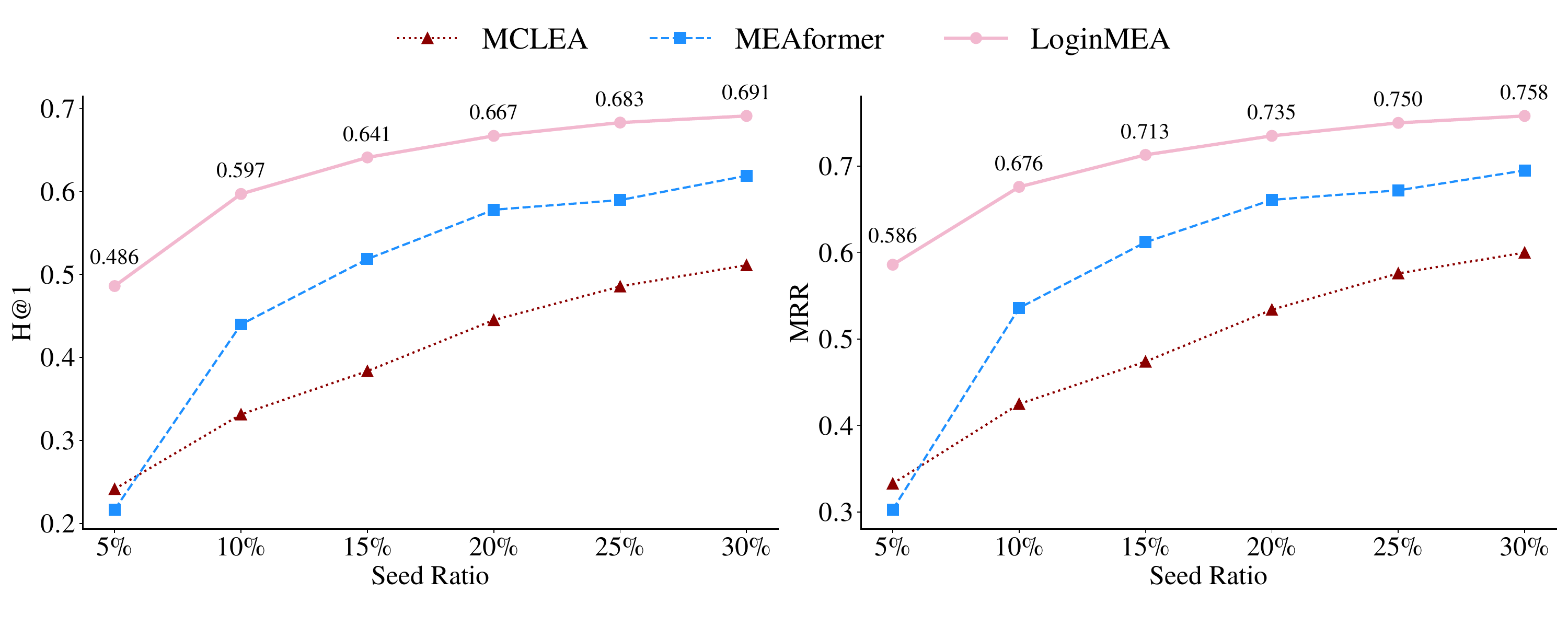}
		\caption{Results in the low-resource data scenario with proportions of seed alignments on FB15K-DB15K dataset.}
		\label{seed_ratio}
	\end{center}
\end{figure}

\subsection{Performance on Low-Resource Training Data}
To further explore the performance with low-resource training data, we vary the alignment seed ratio from $5\%$ to $30\%$.
The latest baselines MCLEA~\cite{MCLEA} and MEAformer~\cite{Meaformer} are compared in Figure~\ref{seed_ratio}. 
We can observe that as the proportion of alignment seeds decreases, the performance of all methods tend to decrease in terms of metrics.
However, it is obvious that our LoginMEA continuously outperforms MCLEA and MEAformer, which indicates the effectiveness of our proposed method especially under the low-resource scenarios. 
Moreover, it is worth noting that the gap between them is much more significant when the seed alignments are extremely few (5\%), 
which guarantees the reliability and robustness of LoginMEA under extremely low-resource scenarios with local-to-global interactions.

\section{Conclusion}
In this paper, we propose a novel local-to-global interaction network for MMEA, termed as LoginMEA, by facilitating the interactions of multi-modal information and relational graph structures. 
Particularly, we develop a local multi-modal interactive fusion module to capture diverse impacts and element-level interactions among modalities.
Besides, we devise a global multi-modal interactive aggregation module to fully capture relational associations between entities with their multi-modal information.
Empirical results show that LoginMEA consistently outperforms competitors across all datasets and metrics.
Further experiments demonstrate the effectiveness of the multi-modal fusion paradigm and the robustness of LoginMEA in low-resource scenarios.



\ack
We would like to thank the anonymous reviewers for their comments. This work was supported by the National Key Research and Development Program of China (Grant No.2021YFB3100600), the Youth Innovation Promotion Association of CAS (No.2021153), and the Postdoctoral Fellowship Program of CPSF (No.GZC20232968).


\bibliography{ecai}
\end{document}